# Physics-Informed Extreme Learning Machine (PIELM): Opportunities and Challenges

He **Yang**[1#], Fei **Ren**[2,5#], Francesco **Calabrò**[3], Hai-Sui **Yu**[4], Xiaohui **Chen**[4], Pei-Zhi **Zhuang**[1*]

1. School of Qilu Transportation, Shandong University
2. School of Mechanical Engineering, Shandong Key Laboratory of CNC Machine Tool Functional Components, Qilu University of Technology (Shandong Academy of Sciences), Jinan 250353, China；
3. Dipartimento di Matematica e Applicazioni, University of Naples Federico II, Napoli, Via Cintia 80126, Italy
4. School of Civil Engineering, University of Leeds, Leeds, LS2 9JT, UK
5. Shandong Institute of Mechanical Design and Research, Jinan 250031, China

*Corresponding author: zhuangpeizhi@sdu.edu.cn; yanghesdu@mail.sdu.edu.cn;

[#]These authors contributed equally to this work.

## Abstract

We are delighted to see the recent development of physics-informed extreme learning machine (PIELM) for its higher computational efficiency and accuracy compared to other physics-informed machine learning (PIML) paradigms. Since a comprehensive summary or review of PIELM is currently unavailable, we would like to take this opportunity to share our perspectives and experiences on this promising research direction. We can see that many efforts have been made to solve ordinary/partial differential equations (ODEs/PDEs) characterized by sharp gradients, nonlinearities, high-frequency behavior, hard constraints, uncertainty, multiphysics coupling, and interpretability. Despite these encouraging successes, many pressing challenges remain to be tackled, which also provides opportunities to develop more robust, interpretable, and generalizable PIELM frameworks for scientific and engineering applications.

# 1. Introduction

The rapid development of physics-informed machine learning (PIML) has presented both challenges and opportunities for science and engineering in recent years (Karniadakis et al. 2021; Toscano et al. 2025). Compared to conventional numerical methods, PIML seamlessly integrates physical laws and monitoring data within neural networks, offering a promising collaborative data-physics driven approach for solving ordinary/partial differential equations (ODEs/PDEs). Among PIML paradigms, the latest five years have frequently seen diverse applications of physics-informed neural networks (PINNs) in scientific research and engineering practice (Xu et al. 2022; Sharma et al. 2023; Ahmadi et al. 2025; Yuan et al. 2025), which are able to solve ODEs/PDEs with or without monitoring data using deep neural network architecture (see Figure 1 a). At the same time, their limitations are also gradually coming into our eyes owing to the inherent architecture of deep neural networks. PINNs often require extensive training times, involve laborious hyperparameter tuning, and suffer from finding loss weights for loss terms. Also, they still face challenges when addressing strongly nonlinear or multiscale problems.

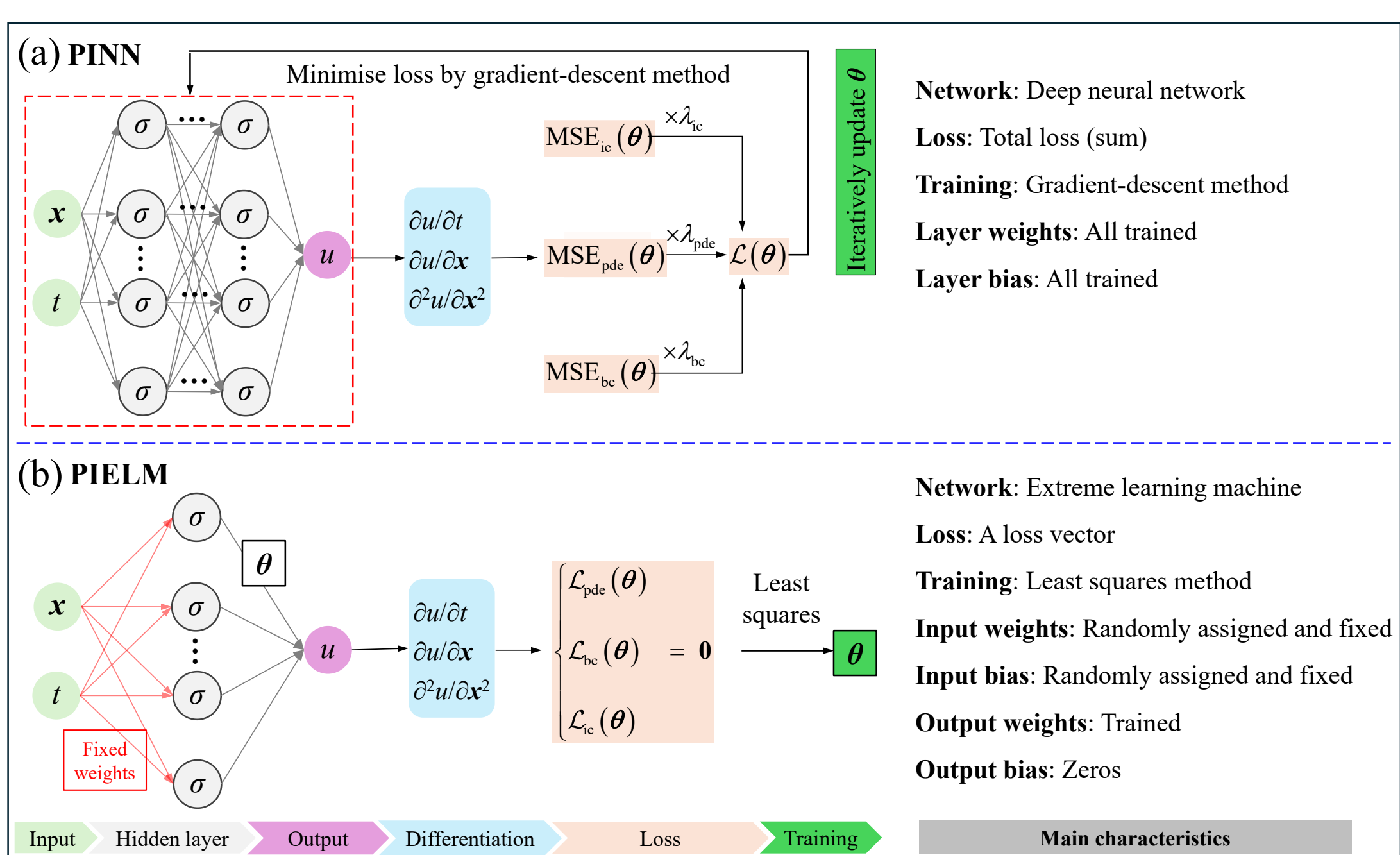

Note: in this figure $\boldsymbol{x}$ and $t$ denote the spatial and temporal variables; $\sigma$ is the activation function; $u$ is the

underdetermined solution; In the PINN framework, $MSE_{pde}$, $MSE_{ic}$ and $MSE_{bc}$ denote the mean square errors related to governing PDEs, initial conditions and boundary conditions, respectively; In the PIELM framework, $\mathcal{L}_{pde}$, $\mathcal{L}_{ic}$ and $\mathcal{L}_{bc}$ are loss vector terms related to governing PDEs, initial conditions and boundary conditions, respectively; $\boldsymbol{\theta}$ is trainable parameter for neural network.

**Figure 1 PINN and PIELM frameworks**

Alongside the development of PINNs, an alternative is the physics-informed extreme learning machine (PIELM) (Dwivedi and Srinivasan 2020a; Calabrò et al. 2021) that solves ODEs/PDEs using extreme learning machine (ELM) network (Huang et al. 2006) instead of deep neural networks. ELM network is a single-layer feed forward network and is trained by the least squares method (see Figure 1 b). Compared to deep neural networks, it can maintain architectural simplicity, significantly enhance training efficiency and accuracy, and reduce hyperparameter number. These inherent characteristics make ELM particularly suitable for PIML to accelerate the training process. For instance, when solving the PDEs for Stefan problems, PIELM has been shown to improve solution accuracy from of $10^{-3}$~$10^{-5}$ to $10^{-6}$~$10^{-8}$ and reduce over 98% training time (Ren et al. 2025b).

On the other hand, the high performance of PIELM does not come without trade-offs. Training PIELM is direct and non-iterative, so PIELM cannot be trained in batches and in certain aspects PIELM faces even greater challenges than PINNs, such as those ODEs/PDEs with sharp gradients, nonlinearities, high-frequency behavior, uncertainty, multiphysics coupling, and interpretability. Fortunately, numerous recent studies have extended the original PIELM framework in various directions to address these issues. As there is currently no comprehensive perspective or review paper on this emerging topic, this perspective paper aims to provide a timely summary and outlook on the recent achievements and challenges of PIELM. The main structure of this perspective is shown in Figure 2, while a list of publications on PIELM is summarized in the Supporting Information.

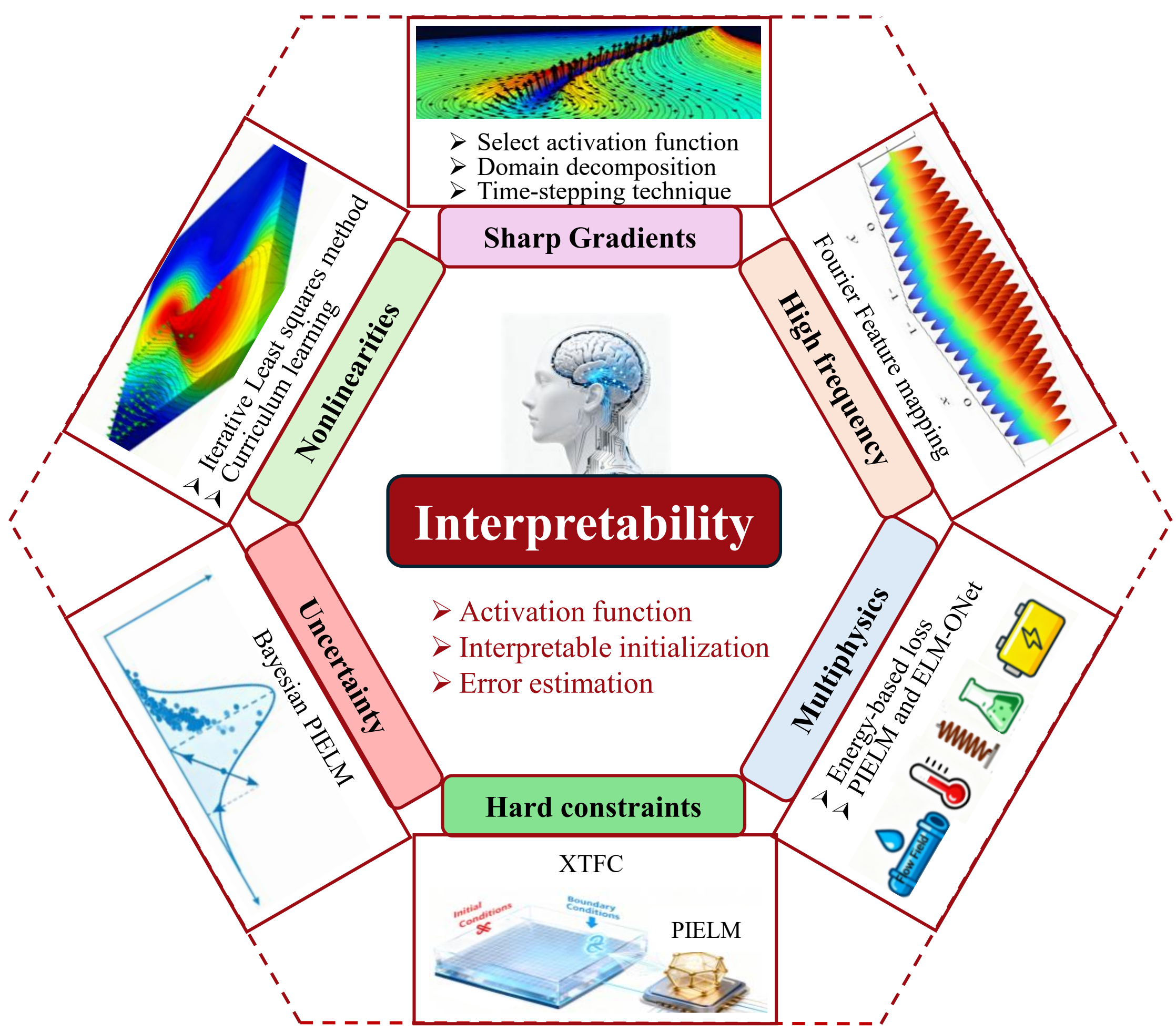

**Figure 2 Structure of the perspective**

## 2. ODEs/PDEs with Sharp gradient

The sharp gradient may be the one of the most challenging problems confusing PIML. More collocation points are normally required to train machine learning models for ODEs/PDEs. Nevertheless, this strategy is not always suitable for PIELM, because increasing training data cannot be input into ELM network batch by batch and numerical instability arises when inverting large matrices. Three main strategies have been proposed to mitigate these issues.

Calabrò et al. (2021) solved several ODEs with sharp gradients by replacing the widely used tanh activation function (i.e., $\sigma(x)=\left(e^{x}-e^{-x}\right)/\left(e^{x}+e^{-x}\right)$) with a sigmoid function (i.e., $\sigma(x)=1/\left[1+\exp(-x)\right]$). Taking several diffusion-advection-reaction problems as examples, they demonstrated that high accuracy was achieved and sometimes outperforms the finite difference method. Although their benchmark cases were restricted to ODEs, they inspired us that selection of activation functions should match the target ODEs/PDEs, and input weights and bias should be properly initialized. However, selecting activation functions and initializing hyperparameters are often problem-dependent and not universally generalizable for ODEs/PDEs with sharp gradients. Unified mathematics and theoretical foundations have not been available so far and are urgent in the future, then this strategy may become the best choice to solve those ODEs/PDEs.

The second strategy is the involvement of solution domain decomposition and local neural networks (or Extend PIELM with different techniques for interfaces) (Dwivedi and Srinivasan 2020a; Dong and Li 2021; Sun and Chen 2025; Zhu et al. 2025). We may separate the whole solution domain into several sub-domains, and in each domain different activations functions may be selected with different initializations to approximate target solution. When this technique is adopted, Dong and Li (2021) demonstrated that the prediction error of PIELM generally decrease exponentially with the increase of sub-domain and collocation points. Simultaneously, the training time will generally increase with the number of sub-domains, which may be the limitation of this strategy for PIELM.

The time-stepping method (or time-marching method) is the third available strategy for handling sharp gradients in time-dependent problems, forming the time-stepping PIELM (TS-PIELM) (Calabro et al. 2023; Yang et al. 2025b). We can divide the temporal domain into multiple intervals and each is handled by a separate ELM network, where the output of one network serves as the input to the next. Compared to original PIELM, TS-PIELM effectively batches collocation data across increasing time steps. When solving sharp-gradient PDEs with discontinuity between initial and boundary conditions, Calabro et al. (2023) and Yang et al. (2025b) demonstrated that TS-PIELM can smooth the gradient by increasing collocation points at the sharp-gradient domain, thereby improving the accuracy. Another advantage of TS-PIELM lies in the improved training efficiency for high-dimensional PDEs, because the dimensions of coefficient matrix and hidden neuron numbers in each ELM network are significantly reduced. It is interesting to note the difference between TS-PIELM and TS-PINN: the time-stepping method may not be so powerful for PINN, primarily because the training efficiency can be rather slow even if using transfer learning technique. Finally, in our opinion, TS-PIELM can be further improved in two parallel ways: (i) Applying TS-PIELM to inverse problems or parameter inference remains challenging; and (ii) the self-adaptive time-stepping schemes are also essential for TS-PIELM. Combining time-stepping method and temporal domain decomposition strategy may be helpful for the former, while theoretical advances for error-estimation in PIELM network are necessary for the latter.

## 3. ODEs/PDEs with nonlinearities

Solving nonlinear ODEs/PDEs presents another challenge for original PIELM, and two strategies have been developed to address PDE nonlinearities. Dong and Li (2021) proposed an iterative PIELM framework for nonlinear PDEs by optimizing the loss vector using iterative least squares methods. After tens of iterations, iterative PIELM can still achieves higher accuracy and efficiency than conventional PINN with the gradient-descent-based training methods. Ren et al. (2025b) utilized iterative PIELM to solve Stefan problems, where the governing PDEs are linear, but the moving boundary leads to the nonlinearity of the Stefan problems. However, iterative PIELM may struggles with nonlinearities and sharp-gradient PDEs (e.g., Allen–Cahn equations and Burgers' equations), and the set of initial values for iterative least squares method were just randomly assigned. Although repeated experimental tests showed robustness with this random method, possible collapse remain a theoretical concern. Dwivedi et al.

(2025b) combined the curriculum learning and PIELM to solve nonlinear Burgers' equation, where they divided the temporal domain into multiple time blocks and used the predictor-corrector process to reduce nonlinear PDEs into quasi-linear ones. Similar to the time-stepping methods, this strategy may also face difficulty for inverse problems. The performance of this strategy remains to be explored in the future, especially for some other equations.

## 4. ODEs/PDEs with high-frequency behavior

Neural networks tend to learn low-frequency features yet neglect high-frequency ones, which is also known as spectral bias (Cao et al. 2019; Rahaman et al. 2019; Xu et al. 2019). To solve PDEs with high frequency, PINNs and PIELM can incorporate Fourier feature mapping (FFM) (Wang et al. 2021; Jin et al. 2024) or adjust network frequency by adding frequency coefficients (Zhang 2023). Such strategies require prior knowledge of the target solution frequency, which is often unavailable for forward problems and most inverse problems. Therefore, tuning hyperparameters of frequency coefficients is inevitable but is incompatible with PIELM for the advantage of high training efficiency, one of the most important strengths of PIELM. Most recently, our group proposed a general Fourier feature PIELM (GFF-PIELM) by incorporating the FFM-based activation function into ELM network (Ren et al. 2025a). This strategy can not only remain the simplification of PIELM framework but also own the capability of FFM to capture high-frequency features. From our perspective, GFF-PIELM will become one of the mainstreams for solving high-frequency and variable-frequency PDEs, and further improvements based on GFF-PIELM are welcome for diverse applications in science and engineering.

## 5. ODEs/PDEs with hard constraints

Hard constraints are widely used in PIML to ensure that initial and/or boundary conditions are satisfied exactly. This is especially beneficial for PIELM to improve solution accuracy (Lu et al. 2021; De Falco et al. 2025). Schiassi et al. (2021b) proposed a hard-constrained PIELM framework called XTFC, which significantly enhances solution accuracy. Later, Dwivedi et al. (2025b) introduced a gated XTFC for ODEs by combining XTFC with soft, learned domain decomposition. Compared to XTFC, their gated XTFC can save 80% collocation points and 66% training time. Wang and Dong (2024) proposed a so-called A-TFC for high-dimensional PDEs using an approximate variant of the theory of functional connections. In order to conveniently apply hard constraints to high-order ODEs, our group formulated a fourth-order ODE into four first-order ODEs and then solved them by PIELM (Guo et al. 2025). The research experience from our group suggests that hard constraints should be applied as much as possible to improve solution accuracy and in some cases to help solve PDEs with sharp gradients (De Florio et al. 2024b). However, hard constraints are not always feasible for high-dimensional PDEs in solution domain with complex geometries. In the future the applications of proper weak-form constraints in PIELM may also be beneficial (De Falco et al. 2025; Kuvakin et al. 2025). For example, De Falco et al. (2025) proposed a least squares with equality constraints ELM (LSE-ELM) by employing a novel collocation-based strategy rather than imposing hard constraints through modifications to the network architecture, so it is suitable for PDEs with various solution domains and boundaries condition.

## 6. Uncertainty quantification and resistance to noisy data

A key advantage of PIELM over conventional numerical methods is the convenient combination of physics laws and monitoring data, making PIELM suitable for real-time monitoring and early warning systems (Guo et al. 2025). In practice, data are often sparsely distributed and contaminated by measurement errors. The uncertainty quantification is particularly significant for PIELM, because the high fitting capability of ELM networks tends to overfit noisy data. In other words, the original PIELM exhibiting poor robustness against noisy data. Liu et al. (2023) proposed a Bayesian

PIELM (B-PIELM) to enhance resistance to data noise, which is able to provide probabilistic prediction and uncertainty quantification for both forward and inverse problems. Through a specific engineering case (e.g., tunneling in civil engineering), Guo et al. (2025) investigated the effect of the number and spatial distribution of monitoring data on the PIELM performance, and demonstrating that placing sensors near the region with sharper solution gradient yields more accurate predictions. As pointed out in Karniadakis et al. (2021) and Toscano et al. (2024), the uncertainties arising from physical laws, measured data, and neural networks are three of principal sources in PIML. Their coupled effects would further complicate uncertainty analysis, and practical applications of PIELM with uncertainty quantification should be examined in the future.

## 7. PDEs for multiphysics coupling

Multiphysics coupling is a common problem in science and engineering where the multiphysics fields (e.g., thermal, mechanical, hydraulic, chemical, and electrical fields) interact with each other. To solve the multiphysics coupling PDEs by PINNs, one may try to first non-dimensionalize the variables and then establish the loss function with the state-of-the-art self-adaptive loss weighting techniques. However, the prediction accuracy remains unsatisfactory with these strategies (Amini et al. 2022), mainly because multiphysics coupling PDEs normally involve variables that differ in dimension and scale by orders of magnitude. Alternatively, we think there may be two potential alternatives to solve the coupled PDEs. One is to formulate the total loss based on a free energy potential (e.g., Gibbs free energy and Helmholtz free energy), so that energy equilibrium and variables consistency across different coupled fields are inherently maintained (Chen and Wang 2025). The other is the use of PIELM because it does not require defining loss weights in different loss terms and offers a more straightforward framework for multi-physics applications. However, both two potential ways may be computationally expensive. Karniadakis et al. (2021) suggested first training PINNs for each PDE and then solving the coupled using deep operator learning, which has been applied to PDEs coupled between two physical fields (Cai et al. 2021; Kobayashi et al. 2025). On the contrary, operator learning such as physics-informed ELM-ONet has not been clear yet.

## 8. Interpretability of PIELM

Although PIELM achieves a certain degree of explainability through embedded physical laws and a simple network architecture, its intrinsic mechanisms remain insufficiently explored. First, selecting activation functions in PIELM is largely based on experience, rather than theoretical matching with the ODE/PDE features. At present, we may just know that the activation functions such as tanh and radial basis functions, perform well in typical cases (e.g., smooth gradient and low frequency). Fourier-based activation functions are more appropriate for high-frequency PDEs, which are equivalent to FFM (Ren et al. 2025a). Obviously, the optimal activation function varies from one ODEs/PDEs to another, and future efforts could be directed toward: (i) When using new activation functions, we could systematically evaluate their suitability and potential failure across different ODEs/PDEs; and (ii) When solving a target ODE/PDE, we could comprehensively test PIELM performance with multiple activation functions. Such analyses would ultimately enable the establishment of a reference list of optimal activation functions for various classes of ODEs/PDEs.

Second, the input weights and bias are randomly assigned in conventional PIELM to ensure a sufficiently rich set of bases for approximating target functions. The random method undoubtedly leads to redundant utilization of hidden neurons, and we may minimize neuron numbers to reduce the computational cost of determining output weights. In 2025 more attention is paid to the physically interpretable initialization strategies for weights and biases. For instance, Ren et al. (2025a) initialized the frequency coefficients for Fourier-based functions from the observed distribution of ELM output weights. Based on the radial basis activation functions, Dwivedi et al.

(2025b) proposed an interpretable initialization method by linking weights and bias to the locations and characteristic scales of Gaussian functions. Peng et al. (2025) proposed a rank-inspired neural network for the initialization of PIELM. Besides, we tested the performance of PIELM using a pair of orthogonal activation functions (e.g., trigonometric functions), which seems to help eliminate the initialization of input bias. However, this replacement may be more suitable for ODEs and extension to PDEs may be directly achieved. From the authors' perspective, an effective physics-guided initialization strategy should be compatible with the selected activation function and the intrinsic characteristics of the underlying ODE/PDE. We should also fully exploit the ELM output weights that naturally encode the relationship between the target solution and specific hidden neurons, so as to further improve the learning efficiency. These considerations highlight again the importance of establishing a solid mathematical framework for PIELM.

As mentioned earlier, error estimation in PIELM is crucial for solving ODEs/PDEs in the absence of benchmark solutions and is especially important in TS-PIELM when incorporating adaptive time-stepping techniques and domain decomposition strategy. However, priori error estimation framework for PIELM is limited. Existing studies may only assess the model accuracy through a posteriori metrics such as the $L_2$ norm of the prediction error (Ren et al. 2025b), and error estimation methods remains an open question.

## 9. Conclusion

In summary, PIELM and its variants represent a powerful and efficient alternative to conventional PINNs and numerical methods, combining the strength of fast training and data-physics driven analysis. Recent studies have attempted to address challenges faced by PIELM, such as solving PDEs with sharp gradients, nonlinearity, high-frequency behavior, hard constraints, uncertainty quantification, multiphysics coupling, and interpretability. It is encouraging to see a substantial step forward in advancing the original PIELM framework, but significant challenges remain. Future research is expected to develop more robust, interpretable, and generalizable PIELM frameworks with better physical consistency and broader applicability in real world. Such developments will enable PIELM to play a central role in real-time scientific computing, real-time monitoring and early warning, intelligent construction, among others, in science and engineering.

**Supporting Information: Summary of publications on PIELM**

| **Authors** | **PIELM framework** | **Remark** |
|---|---|---|
| Dwivedi and Srinivasan (2020a) | Vanilla PIELM and distributed PIELM | May be seen as the first paper for PIELM |
| Dwivedi and Srinivasan (2020b) | Vanilla PIELM | Fourth-order biharmonic equation |
| (Yang et al. 2020) | Legendre improved extreme learning machine | Legendre polynomials are basis functions |
| Calabrò et al. (2021) | Vanilla PIELM | ODEs with sharp gradients |
| Schiassi et al. (2021b) | XTFC | First paper for hard-constrained PIELM |
| Dong and Li (2021) | Iterative PIELM with domain decomposition strategy | Solved nonlinear PDEs |
| Dwivedi et al. (2021) | Distributed PIELM with time-stepping method | Solved nonlinear PDEs |
| Fabiani et al. (2021) | Iterative PIELM | Solved nonlinear PDEs by the Newton method |
| (Liu et al. 2021) | Chebyshev PIELM | Chebyshev polynomials are basis functions |
| (Schiassi et al. 2021a) | XTFC | May be the first paper for PIELM with hard constraints |
| Schiassi et al. (2022b) | XTFC | |
| Schiassi et al. (2022a) | XTFC | |
| Yan et al. (2022) | PIELM with domain decomposition strategy | Application to linear elastic plate and shell structures. |
| Chen et al. (2022) | PIELM with a unity partition method | Application to elasticity problem and Stokes flow problem |
| Ni and Dong (2023) | Multi-layer PIELM | Improved performance |
| Dong and Yang (2022) | Multi-layer PIELM | May perform better than finite element method |

| Dong and Wang (2023) | PIELM with domain decomposition strategy | Three algorithms for inverse problems |
| --- | --- | --- |
| Calabro et al. (2023) | TS-PIELM | Parabolic PDEs |
| Li et al. (2023) | Kernel PIELM | |
| (Liu et al. 2023) | Bayesian PIELM | Uncertainty and resistance to noisy data |
| Wang and Dong (2024) | A-TFC | Approximate variant of the Theory of Functional Connections for applying hard constraints |
| Li et al. (2024) | PIELM with domain decomposition strategy | Fourth-order biharmonic equation; Test various activation functions and initialization methods |
| Harder et al. (2024) | TS-PIELM | Fourth-order nonlinear PDEs |
| De Florio et al. (2024a) | XTFC | |
| De Florio et al. (2024b) | XTFC | PDEs with sharp gradients |
| Thiruthummal et al. (2024) | XTFC | Nonlinear PDEs |
| (Ahmadi Daryakenari et al. 2024) | XTFC | |
| Schiassi et al. (2024) | XTFC | |
| Schiassi et al. (2024) | XTFC | Optimal control problems |
| Dong et al. (2024) | Hierarchical PIELM | |
| Neufeld et al. (2025) | | High-dimensional nonlinear parabolic partial differential equations (PDEs); Error analysis |
| Liang et al. (2025) | Piecewise PIELM | Interface problems |
| Ren et al. (2025b) | Iterative PIELM | Nonlinear Stefan problems with moving boundaries |
| (Zhuang et al. 2025) | Vanilla PIELM | Inverse Stefan problems with moving boundaries |
| Zhu et al. (2025) | Extended PIELM | Application to linear elastic fracture mechanics |
| (Wang et al. 2025) | Vanilla PIELM | Application to linear elastic problems |
| (Peng et al. 2025) | Rank-inspired PIELM | A novel initialization method incorporating a preconditioning stage |
| (Pang et al. 2025) | PIELM | Application to geotechnical |

| | | engineering |
|---|---|---|
| (Mishra et al. 2025) | Eig-PIELM | Hard constraints; linear eigenvalue problems |
| (Kuvakin et al. 2025) | PIELM with weak-form constraints | |
| (Huang et al. 2025) | Vanilla PIELM | Application to waveguides and transmission lines |
| (Kabasi et al. 2025) | Improved PIELM | Finding tensile membrane structures |
| De Falco et al. (2025) | PIELM with weak-form constraints | |
| Dwivedi et al. (2025a) | XTFC | Soft and learned domain decomposition |
| (Dwivedi et al. 2025b) | PIELM | Curriculum Learning for time-dependent and time-independent PDEs ; physically interpretable initialization |
| Guo et al. (2025) | XTFC | Applied hard constraints to high-order ODEs; data and physics-driven application to geotechnical engineering |
| Yang et al. (2025a) | TS-PIELM | Application to geotechnical engineering |
| (Ren et al. 2025a) | GFF-PIELM | Fourier-based activation function for high-frequency PDEs; physically interpretable initialization |

*Note: In each year the papers is not arranged in chronological order.